\documentclass{article}

    \PassOptionsToPackage{numbers, compress}{natbib}

\usepackage[final]{neurips_2023}

\usepackage[utf8]{inputenc} 
\usepackage[T1]{fontenc}    
\usepackage{hyperref}       
\usepackage{url}            
\usepackage{booktabs}       
\usepackage{amsfonts}       
\usepackage{nicefrac}       
\usepackage{microtype}      
\usepackage{xcolor}         

\usepackage{algorithm}

\usepackage{algpseudocode}
\usepackage{graphicx}

\title{Towards Generalizable SER: Soft Labeling and Data Augmentation for Modeling Temporal Emotion Shifts in Large-Scale Multilingual Speech}

\author{
Mohamed Osman$^{1*}$ \quad Tamer Nadeem$^{1*}$ \quad Ghada Khoriba$^2$ \\
$^1$Virginia Commonwealth University \quad $^2$Nile University\\
\texttt{\{osmanmw,tnadeem\}@vcu.edu} \quad \texttt{ghadakhoriba@nu.edu.eg}\\
}

\begin{document}

\maketitle

\vspace{-0.2in}
\begin{abstract}
\vspace{-0.15in}
Recognizing emotions in spoken communication is crucial for advanced human-machine interaction. Current emotion detection methodologies often display biases when applied cross-corpus. To address this, our study amalgamates 16 diverse datasets, resulting in 375 hours of data across languages like English, Chinese, and Japanese. We propose a soft labeling system to capture gradational emotional intensities. Using the Whisper encoder and data augmentation methods inspired by contrastive learning, our method emphasizes the temporal dynamics of emotions. Our validation on four multilingual datasets demonstrates notable zero-shot generalization. We publish our open source model weights and initial promising results after fine-tuning on Hume-Prosody.
\end{abstract}

\vspace{-0.15in}
\section{Introduction}

\vspace{-0.1in}

Emotions significantly influence human communication, affecting both the message's content and its delivery. As human-machine interfaces evolve, machines' capacity to comprehend and react to human emotions emerges as a research imperative. Speech Emotion Recognition (SER) endeavors to decode the emotional subtleties in spoken language.

However, SER's progression has been impeded by the scarcity of varied emotional speech data. Models often excel within their training domain but underperform in diverse real-world situations.\cite{chen2023pre} To combat this, we curate a comprehensive dataset, as tabulated in Table \ref{tab:my-table}, from 16 diverse sources, ensuring a broad spectrum of emotions, accents, languages, and recording conditions. Beyond data accumulation, we adopt a nuanced soft labeling approach, capturing the gradations of emotional intensity within speech. Our end-to-end deep learning framework combines data augmentation and adversarial training, emphasizing the temporal evolution of emotions and setting the stage for next-generation emotion-sensitive applications. Access our open-source model at \url{https://github.com/spaghettiSystems/emotion_whisper/}.

\begin{table}[h!]
    \centering

    \scriptsize

    \begin{tabular}{lrcrrrl}
    \hline
    Dataset     & Samples     & \multicolumn{1}{r}{\# Speakers} & \# Classes & mean\_dur(s) & total\_dur(h) & Languages\\    
    \hline                                                
        CMU\_MOSEI\cite{zadeh2018multi}  & 23258  & 1000                      & 6    & 7.66         & 49.49       & EN\\             
        CREMA-D\cite{Cao2014}     & 7442   & 91                        & 6    & 2.54         & 5.26        & EN\\
        EmoV-DB\cite{adigwe2018emotional}     & 6893   & 4                         & 5    & 4.96         & 9.49        & EN\\
        IEMOCAP\cite{busso2008iemocap}     & 9866   & 10                        & 9    & 4.46         & 12.24       & EN\\
        JL\_CORPUS\cite{James2018}  & 2227   & 4                         & 10   & 2.12         & 1.31        & EN\\
        MSP-PODCAST\cite{lotfian2017building} & 104096 & 1433                      & 8    & 5.74         & 165.85      & EN\\
        MSP\_IMPROV\cite{Busso2017} & 8265   & 12                        & 4    & 4.08         & 9.37        & EN\\
        Proprietary           & 63723  & 290                       & 8    & 2.87         & 50.73       & EN\\
        RAVDESS\cite{Livingstone2018}     & 2452   & 24                        & 8    & 4.09         & 2.79        & EN\\
        SAVEE\cite{jackson2014surrey}       & 480    & 4                         & 7    & 3.84         & 0.51        & EN\\
        TESS\cite{dupuis2010toronto}        & 2800   & 2                         & 7    & 2.06         & 1.6         & EN\\
        VESUS\cite{sager2019}       & 14868  & 10                        & 5    & 1.73         & 7.13        & EN\\
        ESD\cite{zhou2021emotional}         & 34827  & 20                        & 5    & 2.99         & 28.95       & EN, CN\\
        AVSP\cite{tientcheu2020}        & 13618  & 128                       & 12   & 4.61         & 17.43       & \begin{tabular}[c]{@{}l@{}}EN, FR, CN, Other\end{tabular} \\
        OGVC\cite{ogvc}        & 8851   & 13                        & 9    & 1.56         & 3.84        & JP\\
        STUDIES\cite{saito2022studies}     & 5689   & 3                         & 4    & 5.7          & 9.0         & JP\\
        \hline
        EMO-DB\cite{emodbburkhardt2005database}            & 494   & 10    &  6    & 2.8          & 0.38        & DE   \\
        AESDD\cite{aesddvryzas2018speech}            &  604     &  6     &  5    & 4.10             & 0.69            & GR    \\
        URDU\cite{urdulatif2018cross}            &  400     &   38    &  4    & 2.50             & 0.28            & PK    \\
        MASC\cite{wu2006masc}             &  25636     &   68    &   5   & 1.96             & 13.94            & CN    \\

    \end{tabular}
    \vspace{0.1in}
    \caption{Datasets used in the training and testing of this paper. We present the number of samples, speakers, classes, languages, average duration in seconds, and total duration in hours.}
    \label{tab:my-table}
    \vspace{-0.1in}
    \end{table}

\section{Methodology}
\vspace{-0.1in}

\begin{figure}
    \centering
    
    \makebox[\textwidth][c]{\includegraphics[width=1.2\textwidth]{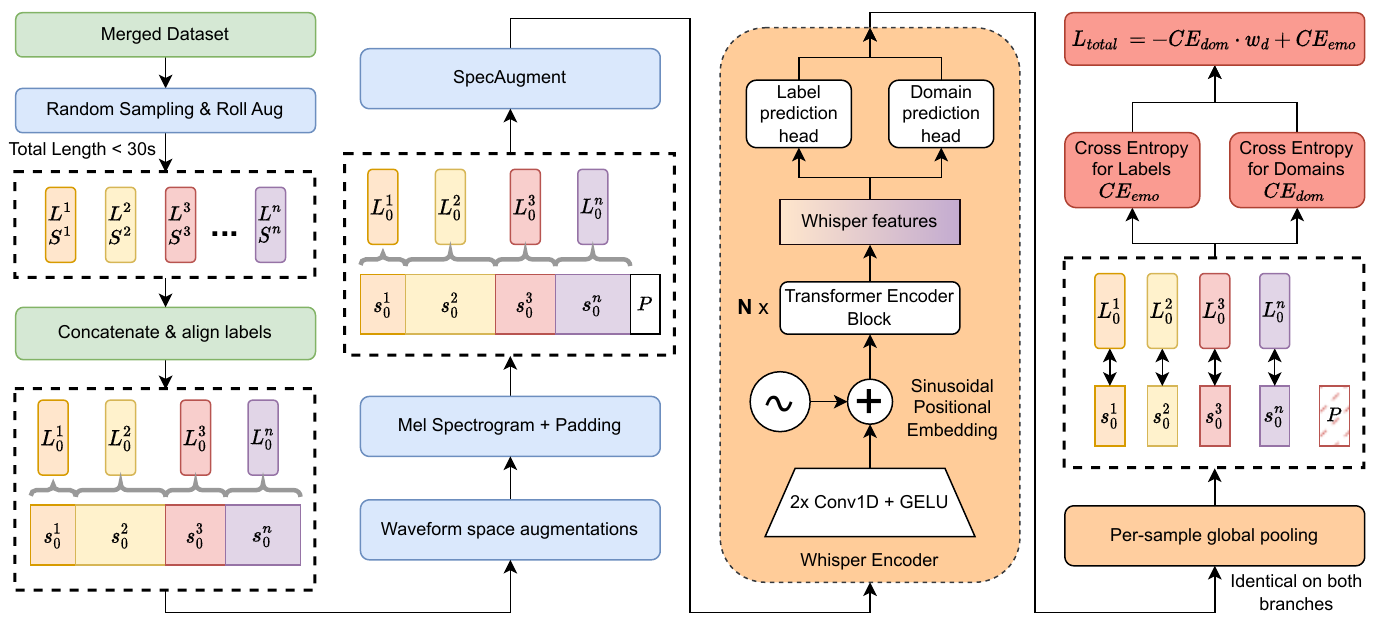}}
    \vspace{-0.15in}
    \caption{System diagram for training, samples concatenation, augmentation, model, and losses computation.}
    \label{fig:methodology}
 
\end{figure}

Figure~\ref{fig:methodology} shows an overview of the training setup we used for our models. In the following subsections, we describe the main components of our methodology.

\vspace{-0.05in}
\subsection{Dataset Sources and Preparation}
\vspace{-0.1in}

Our dataset includes soft labels for the following emotions: 'happiness', 'sadness', 'disgust', 'fear', 'surprise', 'anger', 'other', and 'neutral'. For enhanced generalization, we've added a 'domain' attribute by combining the 'dataset', 'speaker', and 'language' details of each audio. Recognizing the varied labeling across datasets, we've harmonized them using a strategy inspired by \cite{shaver1987emotion}. For example, 'contempt' is mapped to 'disgust' with a weight of 0.5. We use 95\% of our data for training and 5\% for model validation. IEMOCAP's fourth session serves as our test set, and we further evaluate our model's adaptability on unseen out-of-distribution datasets like EmoDB, AESDD, URDU, and MASC.

\vspace{-0.05in}
\subsection{Neutral Smoothing}
\vspace{-0.1in}
While analyzing our datasets, we noticed variations in the available annotations; for instance, RAVDESS includes emotion strength, while CMU-MOSEI and MSP-Podcast offer more detailed voter information. We observed that many samples had low or even no discernible emotional intensity, while other samples had mixed votes, all denoting low strength. We devised a technique to clarify this within the emotional labels of such samples by boosting the weight of the "neutral" emotion. The underlying assumption is that samples with low emotional intensity are more neutral.

Given a dataset, the average emotional intensity, \( \bar{E} \), is calculated as:

\vspace{-0.1in}
\[
\bar{E} = \frac{1}{N} \sum_{i=1}^{N} E_i
\]
\vspace{-0.1in}

where \( N \) is the total number of samples and \( E_i \) is the emotional intensity of the \( i^{th} \) sample, computed as:

\vspace{-0.15in}
\[
E_i = \sum_{j} e_{i,j}
\]
\vspace{-0.1in}

Here, \( e_{i,j} \) represents the emotion score for emotion \( j \) in sample \( i \). If a sample's intensity, \( E_i \), is less than \( \bar{E} \), a neutral smoothing factor, \( \alpha \), is computed:

\vspace{-0.05in}
\[
\alpha = \min\left(\frac{\left|\bar{E} - E_i\right|}{\bar{E}}, 0.45\right)
\]
\vspace{-0.15in}

The updated emotion scores, \( e' \), are:

\vspace{-0.1in}
\[
e'_{j} = e_j(1 - \alpha) + (1 - e_j) \frac{\alpha}{M}
\]
\vspace{-0.15in}

with \( M \) being the total number of emotion categories. The primary emotion is retained by ensuring the maximal score in \( e \) and \( e' \) remains the same. If not, the neutral emotion's weight in \( e' \) is adjusted accordingly.

\vspace{-0.05in}
\subsection{Data Augmentation}
\vspace{-0.1in}
A cornerstone of our methodology is the aggressive on-the-fly audio augmentation strategy. This not only enriches our dataset but also fosters model robustness, which is particularly vital for the temporal prediction of emotional shifts. Our augmentation suite comprises Polarity Inversion, Gain Modification, Audio Reversal, Noise Addition, Resampling, Equalization, and Echo, all applied at an independent random probability of 20\%. In parallel, we also deploy spectrogram-centric augmentations based on SpecAugment \cite{park2019specaugment}, encompassing frequency masking, time masking, and random noise infusion. The rolling part of SpecAugment is performed at the individual waveform level.

\vspace{-0.05in}
\subsection{Data Sampling}
\vspace{-0.1in}
We define our data sampling algorithm in Algorithm \ref{tab:my_algo}. In a nutshell, we sort the dataset by sample length, get the list of samples within our remaining duration (initialized to 30 seconds), randomly select one, remove it from the pool, and repeat until we reach some threshold.

\begin{algorithm}
\caption{Random Sample Picker Algorithm}
\label{tab:my_algo}
\scriptsize
\begin{algorithmic}[1]
\Procedure{PrepareDataset}{$data, start, end$}
    \State Extract subset from $data$ between $start$ and $end$
    \State Sort subset by audio durations
\EndProcedure

\Function{RetrieveSequence}{$L$}
    \State $cumulative\_duration \gets 0$
    \While{$cumulative\_duration < 0.8 \times L$}
        \State Calculate remaining allowable duration as $L - cumulative\_duration$
        \State Using bisect on sorted durations, find set $S$ of samples fitting within the remaining duration
        \If{size of $S$ is small at start}
            \State Refresh the sorted list of samples
            \State Continue
        \EndIf
        \State Randomly select a sample $s$ from $S$
        \State Increment $cumulative\_duration$ by duration of $s$
        \State Remove $s$ from consideration
    \EndWhile
    \State \Return Constructed sequence
\EndFunction
\end{algorithmic}
\end{algorithm}
\vspace{-0.15in}

\subsection{Model Architecture}
\vspace{-0.1in}
We adopt the Whisper-medium model\cite{whisperradford2023robust} for our experiments. We take only the encoder and freeze the convolutional layers. The Whisper encoder is a standard transformer encoder prepended with convolutional layers that increase the dimensionality and cut the sequence length by a factor of two. We add prediction heads for our tasks composed of a single linear layer. Our feature extraction routine is identical to the one Whisper uses. We refer the reader to their paper for further details.

\vspace{-0.05in}
\subsection{Training}
\vspace{-0.1in}
We use multi-class cross-entropy loss for all of our tasks. We define $CE_{emo}$ to be the cross-entropy loss for emotion labels and $CE_{dom}$ to be the cross-entropy loss for the domain labels. We set our final loss as follows $L_{total} = -CE_{dom} * w_d + CE_{emo}$ and find that $w_d = 0.01$ generally works well. We perform gradient \textit{ascent} on the domain loss because we wish to minimize the amount of speaker/domain information that our model recognizes. This is similar to domain-adversarial training\cite{ganin2016domain} except that we conduct no unsupervised learning on any target domains. We use the AdamW~\cite{kingma2014adam, loshchilov2017decoupled} optimizer and Cosine One Cycle\cite{smith2019super} learning schedule. We use gradient accumulation to schedule the batch size to increase over time. We conduct all experiments on 4x A100 SXM4 80GB machines using PyTorch. 

\section{Results}
\vspace{-0.1in}
We present both zero-shot results and adaptation to Hume-Prosody~\cite{schuller2023acm}. All metrics are calculated with only one sample padded to the model's context length (30s) and no masking applied.

\vspace{-0.05in}
\subsection{In Domain \& Zero-shot}
\vspace{-0.1in}
We report the micro F1 metric for all datasets in Table~\ref{tab:zeroshots}. This table also details our model's F1 score for each emotion across all datasets, providing a comprehensive view of its performance. For any of those datasets that do not have labels that our model predicts, we "ban" those labels by adding $-1e27$ to their logits. Further, we use logit adjustment~\cite{menon2020long} based on the distribution of our training set. Datasets completely unseen during training (zero-shot) are denoted as \texttt{Z-<dataset>}; otherwise, they are in-domain as described above. SER models are known to perform poorly under the zero-shot scenario~\cite{chen2023pre}, which limits the scope of their deployment. We show that our model generalizes well even to unseen languages and domains. As there is no directly comparable model that we are aware of, we create `Random F1' which is a Whisper model with a randomly initialized head.

\begin{table}[h!]

    \centering

    \begin{tabular}{ccccccc}
    \hline   
             & Validation & IEMOCAP S4 & Z-EmoDB & Z-MASC & Z-URDU & Z-AESDD \\
        \hline
         Random F1 & 15.25 & 14.35 & 12.15 & 17.64 & 24.75 & 18.05 \\
         Micro F1 & 63.71 & 57.56 & 70.09 & 45.20 & 52.49 & 61.92 \\
         Anger & 77.01 & 70.76 & 86.92 & 48.61 & 72.09 & 74.36 \\
         Disgust & 27.11 & - & 55.17 & - & - & 39.71 \\
         Fear & 74.48 & - & 76.80 & 32.00 & - & 63.44 \\
         Happiness & 74.10 & 52.88 & 69.63 & 37.72 & 55.81 & 63.78 \\
         Neutral & 62.03 & 24.38 & 52.83 & 46.71 & 33.33 & - \\
         Other & 64.71 & - & - & - & - & - \\
         Sadness & 72.35 & 61.72 & 70.80 & 54.96 & 36.60 & 71.59 \\
         Surprise & 64.11 & 24.00 & - & - & - & - \\
    \end{tabular}
    \caption{Micro F1 score on testing datasets under zero-shot setting}
    \label{tab:zeroshots}
\vspace{-0.2in}
\end{table}

\vspace{-0.05in}
\subsection{Hume-Prosody}
\vspace{-0.1in}
 For this task, we adapt our model by removing the existing prediction heads and randomly initializing a new one with nine classes. We fine-tune the training portion of the data with the same methodology defined above, but we employ the mean squared error metric and sigmoid function instead of cross entropy. We present the result of our first training run, which involved no hyperparameter tuning yet outperformed the strong baseline provided in the competition, in Table~\ref{tab:hume}. We report Pearsons' Correlation Coefficient across the mean of the classes and micro F1 score. We were not provided with annotations for the test portion of the data; therefore, we are unable to provide comparison on that.

\begin{table}[h!]
    \centering

    \begin{tabular}{ccc}
    \hline   
        Method & Corr on Dev & F1\\
        \hline   
        \textbf{Ours} & \textbf{0.511} & \textbf{78.83}\\
        Wav2Vec2 & 0.500 & - \\
         ComParE & 0.359 & - \\
         Late Fusion & 0.470 & - \\
    \end{tabular}
    \caption{Correlation and F1 score compared to the baseline method. Our method presented in bold.}
    \label{tab:hume}
\vspace{-0.3in}
\end{table}

\section{Conclusion}
\vspace{-0.1in}
In this paper, we present our methodology and provide very early results on multiple datasets. Remarkably, the model generalizes well on out-of-distribution data under the zero-shot setting. 
Further, we were surprised to exceed the performance of the strong baseline with only one attempt. In the future, we hope to extend this work with thorough evaluations of out-of-distribution datasets under the zero-shot and few-shot settings. In our testing, we observed strong performance in predicting emotion changes temporally, but we are still developing a strong test suite to evaluate how well the model predicts temporal changes in emotion.

\section{Acknowledgements}
High Performance Computing resources provided by the High Performance Research Computing (HPRC) core facility at Virginia Commonwealth University (https://hprc.vcu.edu) were used for conducting the research reported in this work. Special thanks to Professor Alberto Cano for providing these resources on short notice. Further, we would like to thank StabilityAI for providing computational resources at the start of this project.

\vspace{-0.1in}
{\small
\bibliographystyle{IEEEtran} 
\bibliography{references}

\begin{thebibliography}{10}
\providecommand{\url}[1]{#1}
\csname url@samestyle\endcsname
\providecommand{\newblock}{\relax}
\providecommand{\bibinfo}[2]{#2}
\providecommand{\BIBentrySTDinterwordspacing}{\spaceskip=0pt\relax}
\providecommand{\BIBentryALTinterwordstretchfactor}{4}
\providecommand{\BIBentryALTinterwordspacing}{\spaceskip=\fontdimen2\font plus
\BIBentryALTinterwordstretchfactor\fontdimen3\font minus \fontdimen4\font\relax}
\providecommand{\BIBforeignlanguage}[2]{{%
\expandafter\ifx\csname l@#1\endcsname\relax
\typeout{** WARNING: IEEEtran.bst: No hyphenation pattern has been}%
\typeout{** loaded for the language `#1'. Using the pattern for}%
\typeout{** the default language instead.}%
\else
\language=\csname l@#1\endcsname
\fi
#2}}
\providecommand{\BIBdecl}{\relax}
\BIBdecl

\bibitem{chen2023pre}
M.~Chen and Z.~Yu, ``Pre-finetuning for few-shot emotional speech recognition,'' \emph{arXiv preprint arXiv:2302.12921}, 2023.

\bibitem{zadeh2018multi}
A.~Zadeh, P.~P. Liang, S.~Poria, P.~Vij, E.~Cambria, and L.-P. Morency, ``Multi-attention recurrent network for human communication comprehension,'' in \emph{Thirty-Second AAAI Conference on Artificial Intelligence}, 2018.

\bibitem{Cao2014}
H.~Cao, D.~G. Cooper, M.~K. Keutmann, R.~C. Gur, A.~Nenkova, and R.~Verma, ``Crema-d: Crowd-sourced emotional multimodal actors dataset,'' \emph{IEEE Transactions on Affective Computing}, vol.~5, pp. 377--390, 10 2014.

\bibitem{adigwe2018emotional}
A.~Adigwe, N.~Tits, K.~E. Haddad, S.~Ostadabbas, and T.~Dutoit, ``The emotional voices database: Towards controlling the emotion dimension in voice generation systems,'' \emph{arXiv preprint arXiv:1806.09514}, 2018.

\bibitem{busso2008iemocap}
C.~Busso, M.~Bulut, C.-C. Lee, A.~Kazemzadeh, E.~Mower, S.~Kim, J.~N. Chang, S.~Lee, and S.~S. Narayanan, ``Iemocap: Interactive emotional dyadic motion capture database,'' \emph{Language resources and evaluation}, vol.~42, pp. 335--359, 2008.

\bibitem{James2018}
J.~James, L.~Tian, and C.~I. Watson, ``An open source emotional speech corpus for human robot interaction applications,'' vol. 2018-September.\hskip 1em plus 0.5em minus 0.4em\relax International Speech Communication Association, 2018, pp. 2768--2772.

\bibitem{lotfian2017building}
R.~Lotfian and C.~Busso, ``Building naturalistic emotionally balanced speech corpus by retrieving emotional speech from existing podcast recordings,'' \emph{IEEE Transactions on Affective Computing}, vol.~10, no.~4, pp. 471--483, 2017.

\bibitem{Busso2017}
C.~Busso, S.~Parthasarathy, A.~Burmania, M.~Abdelwahab, N.~Sadoughi, and E.~M. Provost, ``Msp-improv: An acted corpus of dyadic interactions to study emotion perception,'' \emph{IEEE Transactions on Affective Computing}, vol.~8, pp. 67--80, 1 2017.

\bibitem{Livingstone2018}
\BIBentryALTinterwordspacing
S.~R. Livingstone and F.~A. Russo, ``The ryerson audio-visual database of emotional speech and song (ravdess): A dynamic, multimodal set of facial and vocal expressions in north american english,'' 2018. [Online]. Available: \url{https://www.}
\BIBentrySTDinterwordspacing

\bibitem{jackson2014surrey}
P.~Jackson and S.~Haq, ``Surrey audio-visual expressed emotion (savee) database,'' \emph{University of Surrey: Guildford, UK}, 2014.

\bibitem{dupuis2010toronto}
K.~Dupuis and M.~K. Pichora-Fuller, ``Toronto emotional speech set (tess)-younger talker\_happy,'' 2010.

\bibitem{sager2019}
J.~Sager, R.~Shankar, J.~Reinhold, and A.~Venkataraman, ``Vesus: A crowd-annotated database to study emotion production and perception in spoken english,'' vol. 2019-September.\hskip 1em plus 0.5em minus 0.4em\relax International Speech Communication Association, 2019, pp. 316--320.

\bibitem{zhou2021emotional}
``Emotional voice conversion: Theory, databases and esd,'' \emph{Speech Communication}, vol. 137, pp. 1--18, 2022.

\bibitem{tientcheu2020}
\BIBentryALTinterwordspacing
D.~Tientcheu, T.~Landry, Q.~He, H.~Yan, and Y.~Li, ``Asvp-esd: A dataset and its benchmark for emotion recognition using both speech and non-speech utterances,'' 2020. [Online]. Available: \url{https://zenodo.org/record/3782416}
\BIBentrySTDinterwordspacing

\bibitem{ogvc}
\BIBentryALTinterwordspacing
 [Online]. Available: \url{https://research.nii.ac.jp/src/en/OGVC.html}
\BIBentrySTDinterwordspacing

\bibitem{saito2022studies}
Y.~Saito, Y.~Nishimura, S.~Takamichi, K.~Tachibana, and H.~Saruwatari, ``Studies: Corpus of japanese empathetic dialogue speech towards friendly voice agent,'' \emph{arXiv preprint arXiv:2203.14757}, 2022.

\bibitem{emodbburkhardt2005database}
F.~Burkhardt, A.~Paeschke, M.~Rolfes, W.~F. Sendlmeier, B.~Weiss \emph{et~al.}, ``A database of german emotional speech.'' in \emph{Interspeech}, vol.~5, 2005, pp. 1517--1520.

\bibitem{aesddvryzas2018speech}
N.~Vryzas, R.~Kotsakis, A.~Liatsou, C.~A. Dimoulas, and G.~Kalliris, ``Speech emotion recognition for performance interaction,'' \emph{Journal of the Audio Engineering Society}, vol.~66, no.~6, pp. 457--467, 2018.

\bibitem{urdulatif2018cross}
S.~Latif, A.~Qayyum, M.~Usman, and J.~Qadir, ``Cross lingual speech emotion recognition: Urdu vs. western languages,'' in \emph{2018 International conference on frontiers of information technology (FIT)}.\hskip 1em plus 0.5em minus 0.4em\relax IEEE, 2018, pp. 88--93.

\bibitem{wu2006masc}
T.~Wu, Y.~Yang, Z.~Wu, and D.~Li, ``Masc: A speech corpus in mandarin for emotion analysis and affective speaker recognition,'' in \emph{2006 IEEE Odyssey-the speaker and language recognition workshop}.\hskip 1em plus 0.5em minus 0.4em\relax IEEE, 2006, pp. 1--5.

\bibitem{shaver1987emotion}
P.~Shaver, J.~Schwartz, D.~Kirson, and C.~O'connor, ``Emotion knowledge: further exploration of a prototype approach.'' \emph{Journal of personality and social psychology}, vol.~52, no.~6, p. 1061, 1987.

\bibitem{park2019specaugment}
D.~S. Park, W.~Chan, Y.~Zhang, C.-C. Chiu, B.~Zoph, E.~D. Cubuk, and Q.~V. Le, ``Specaugment: A simple data augmentation method for automatic speech recognition,'' \emph{Interspeech 2019}, 2019.

\bibitem{whisperradford2023robust}
A.~Radford, J.~W. Kim, T.~Xu, G.~Brockman, C.~McLeavey, and I.~Sutskever, ``Robust speech recognition via large-scale weak supervision,'' in \emph{International Conference on Machine Learning}.\hskip 1em plus 0.5em minus 0.4em\relax PMLR, 2023, pp. 28\,492--28\,518.

\bibitem{ganin2016domain}
Y.~Ganin, E.~Ustinova, H.~Ajakan, P.~Germain, H.~Larochelle, F.~Laviolette, M.~Marchand, and V.~Lempitsky, ``Domain-adversarial training of neural networks,'' \emph{The journal of machine learning research}, vol.~17, no.~1, pp. 2096--2030, 2016.

\bibitem{kingma2014adam}
D.~P. Kingma and J.~Ba, ``Adam: A method for stochastic optimization,'' \emph{arXiv preprint arXiv:1412.6980}, 2014.

\bibitem{loshchilov2017decoupled}
I.~Loshchilov and F.~Hutter, ``Decoupled weight decay regularization,'' \emph{arXiv preprint arXiv:1711.05101}, 2017.

\bibitem{smith2019super}
L.~N. Smith and N.~Topin, ``Super-convergence: Very fast training of neural networks using large learning rates,'' in \emph{Artificial intelligence and machine learning for multi-domain operations applications}, vol. 11006.\hskip 1em plus 0.5em minus 0.4em\relax SPIE, 2019, pp. 369--386.

\bibitem{schuller2023acm}
B.~W. Schuller, A.~Batliner, S.~Amiriparian, A.~Barnhill, M.~Gerczuk, A.~Triantafyllopoulos, A.~Baird, P.~Tzirakis, C.~Gagne, A.~S. Cowen \emph{et~al.}, ``The acm multimedia 2023 computational paralinguistics challenge: Emotion share \& requests,'' \emph{arXiv preprint arXiv:2304.14882}, 2023.

\bibitem{menon2020long}
A.~K. Menon, S.~Jayasumana, A.~S. Rawat, H.~Jain, A.~Veit, and S.~Kumar, ``Long-tail learning via logit adjustment,'' \emph{arXiv preprint arXiv:2007.07314}, 2020.

\end{thebibliography}
}

\end{document}